\documentclass[final,5p,times,twocolumn,authoryear]{elsarticle}

\usepackage{amssymb}
\usepackage{amsmath}
\usepackage{booktabs}
\usepackage{xcolor}
\usepackage{fancyhdr}
\journal{Elsevier}

\begin{document}

\begin{frontmatter}

%% Title, authors and addresses

%% use the tnoteref command within \title for footnotes;
%% use the tnotetext command for theassociated footnote;
%% use the fnref command within \author or \affiliation for footnotes;
%% use the fntext command for theassociated footnote;
%% use the corref command within \author for corresponding author footnotes;
%% use the cortext command for theassociated footnote;
%% use the ead command for the email address,
%% and the form \ead[url] for the home page:
\title{3D Reconstruction and Information Fusion between Dormant and Canopy Seasons in Commercial Orchards Using Deep Learning and Fast GICP}
%% Authors
%% Authors with correct footnote and correspondence
\author[a]{Ranjan Sapkota\fnref{equal}}
\author[b]{Zhichao Meng\fnref{equal}}
\author[a]{Martin Churuvija}
\author[b]{Xiaoqiang Du}
\author[b]{Zenghong Ma}
\author[a]{Manoj Karkee\corref{cor1}}

%% Author affiliations
\affiliation[a]{organization={Department of Biological \& Environmental Engineering, Cornell University},%
            addressline={Riley-Robb Hall, 106, 111 Wing Dr}, 
            city={Ithaca},
            postcode={14850}, 
            state={New York},
            country={USA}}

\affiliation[b]{organization={School of Mechanical Engineering, Zhejiang Sci-Tech University},%
            addressline={Hangzhou 310018}, 
            city={Hangzhou},
            country={China}}
%% Footnotes
\fntext[equal]{These authors contributed equally to this work and share first authorship.}
\cortext[cor1]{Corresponding author: Manoj Karkee, Email: mk2684@cornell.edu}

%% Abstract
\begin{abstract}
In orchard automation, dense foliage during the canopy season severely occludes tree structures, minimizing visibility to various canopy parts such as trunks and branches, which limits the ability of a machine vision system in estimating crucial canopy parameters such as trunk diameter and branch spacing. However, the canopy structure is more open and visible during the dormant season when trees are defoliated. In this work, we present an information fusion framework that integrates multi-seasonal structural data to support robotic and automated crop load management during the entire growing season. The framework combines high-resolution RGB-D imagery from both dormant and canopy periods using YOLOv9-Seg for instance segmentation, Kinect Fusion for 3D reconstruction, and Fast Generalized Iterative Closest Point (Fast GICP) for model alignment. Segmentation outputs from YOLOv9-Seg were used to extract depth-informed masks, which enabled accurate 3D point cloud reconstruction via Kinect Fusion; these reconstructed models from each season were subsequently aligned using Fast GICP to achieve spatially coherent multi-season fusion. The YOLOv9-Seg model, trained on manually annotated images, achieved a mean squared error (MSE) of 0.0047 and segmentation mAP@50 scores up to 0.78 for trunks in dormant season dataset. Kinect Fusion enabled accurate reconstruction of tree geometry, validated with field measurements resulting in root mean square errors (RMSE) of 5.23 mm for trunk diameter, 4.50 mm for branch diameter, and 13.72 mm for branch spacing. Fast GICP achieved precise cross-seasonal registration with a minimum fitness score of 0.00197, allowing integrated, comprehensive tree structure modeling despite heavy occlusions during the growing season. This fused structural representation enables robotic systems to access otherwise obscured architectural information, improving the precision of pruning, thinning, and other automated orchard operations. By bridging the seasonal visibility gap, our methodology advances year-round perception in orchard environments and provides a validated foundation for advancing agricultural automation and robotics.
\end{abstract}

%% Keywords

\end{frontmatter}

%% Add \usepackage{lineno} before \begin{document} and uncomment 
%% following line to enable line numbers
%% \linenumbers

%% main text
%%

%% Use \section commands to start a section
\section{Introduction}
\begin{figure*}[h!]
    \centering
    \includegraphics[width=6.9 in, height=6.9 in]{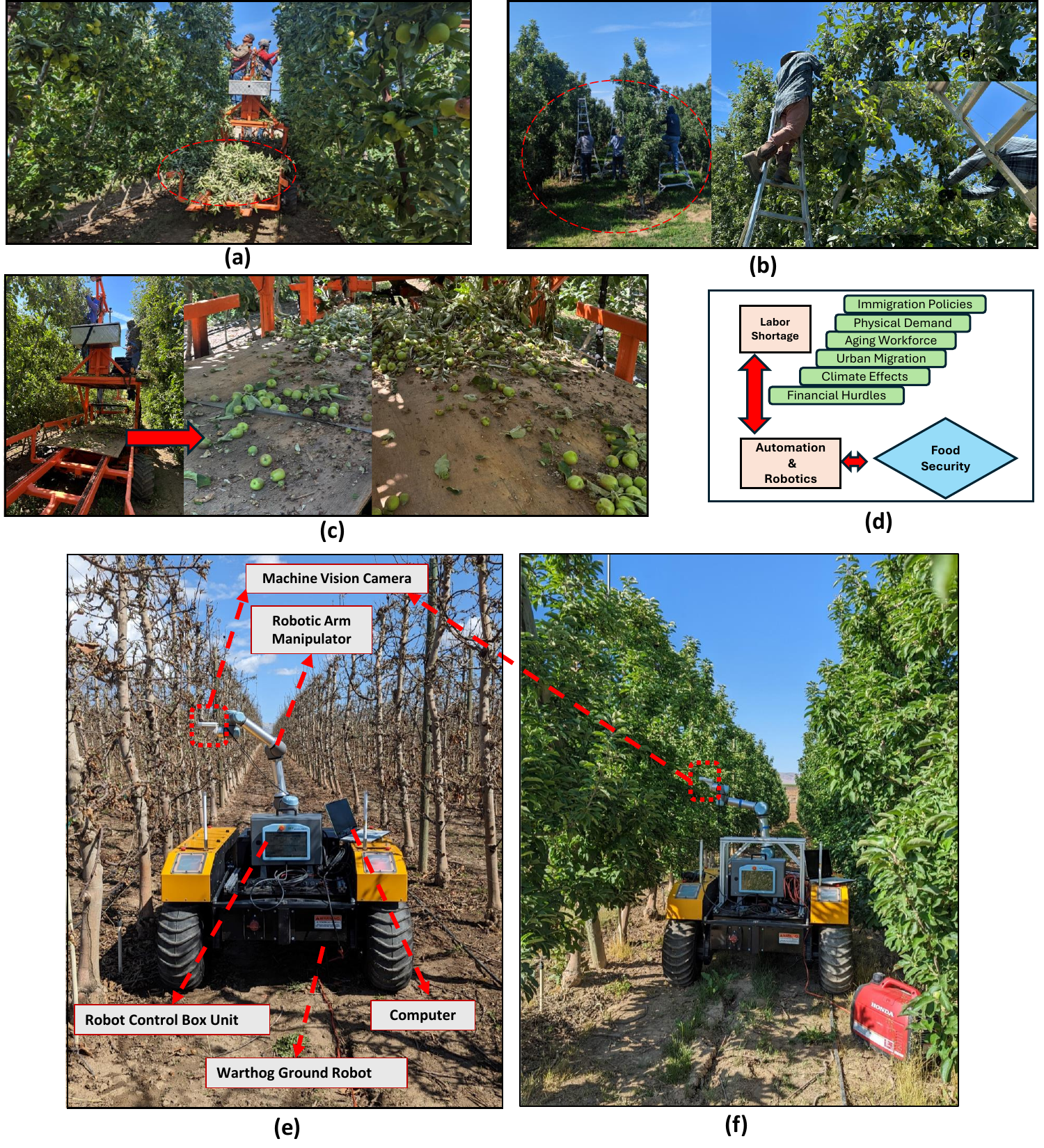}
    \caption{Overview: (a) Workers use a mechanized platform for pruning, enhancing sunlight and airflow, with pruned branches marked by a red circle; (b) Manual thinning with ladders, highlighted by red circles for demanding tasks; (c) Efficient tree access via ground vehicle; (d) Labor challenges; (e) Robotic platform captures tree data in dormant (December) and (f) canopy seasons (June) at a Washington State orchard.}
    \label{fig:Figure1}
\end{figure*}

One of the most labor-intensive operations in commercial orchards occurs during the canopy season  (Figure \ref{fig:Figure1}), a period characterized by dense foliage cover that significantly complicates manual labor tasks \cite{miller2011performance}. Despite advances in agricultural technology, automation and robotics have yet to develop basic functionalities to operate effectively in such complex environments where trees are heavily covered by leaves and canopy foliage. Figure \ref{fig:Figure1} illustrates the labor-intensive nature of crop load management in commercial orchards, showcasing two primary operations observed in Washington State. Figure \ref{fig:Figure1}a depicts manual workers on a mobile platform, pruning tree branches to enhance sunlight penetration and airflow, essential for optimal fruit development; a red dotted circle highlights the branches pruned by professional orchard workers. Likewise, Figure \ref{fig:Figure1}b captures the green fruit thinning operation, where workers (manually) manually carry aluminum ladders from tree to tree, climbing up and down to thin heavily clustered immature fruits, ensuring the remaining fruits develop to meet market size and quality standards. This image also shows the complex and physically demanding body movements involved in this task. Figure \ref{fig:Figure1}c illustrates an alternative strategy for fruitlet thinning using a mobile platform that allows navigation between rows, enabling workers to reach various tree heights more efficiently; the left and middle images display the immature fruitlets removed by the workers in June, in Washington State, USA.

 However, the agricultural sector in the United States, particularly within orchards, is experiencing a significant labor shortage that poses a threat to both domestic and global food security \cite{christiaensen2021future, weiler2022seeing}. This shortage is primarily driven by a trend of urban migration \cite{klocker2020exploring}, where fewer individuals are available or willing to undertake the demanding tasks required in orchard management, such as pruning, thinning, and harvesting \cite{polat2022orchard, zhang2021agricultural}. The situation has been further exacerbated by the post-COVID economic landscape \cite{laborde2020covid, tougeron2021impact}, which has seen a shift in workforce dynamics and an increased reluctance towards physically demanding jobs. Moreover, the aging demographic of farmers \cite{fried2016aging}, coupled with a lack of interest from younger generations in pursuing agricultural careers \cite{prasetyaningrum2022perception}, due to perceived low prestige and long hours \cite{prasetyaningrum2022perception, usman2021strained}, is leading to a workforce gap. This method involves frequent climbing and carrying heavy ladders, leading to extensive physical exertion. Unfortunately, such labor-intensive practices are associated with a high incidence of spinal injuries among workers \cite{gao2022study, fathallah2010musculoskeletal, lee2021prevalence}, significantly impacting their long-term health and quality of life.

As illustrated in Figure \ref{fig:Figure1}d, to address the critical workforce shortages in orchards and to sustainably feed a growing global population, the adoption of advanced automation and robotics emerges as the essential solutions to labor-intensive orchard operations. However, their effectiveness during the canopy season is severely restricted by the orchard's complex structure and the concealment of crucial tree structures by dense foliage. Key architectural elements like branches remain hidden under leaves, significantly challenging the development of automated systems. Notably, over the past two decades, research in orchard automation has predominantly focused on harvesting \cite{zhang2016development, mhamed2024advances}, with limited exploration into robotic or automated systems for other critical tasks such as pruning and thinning. 

To date, the development of fully automated robots capable of operating effectively during the canopy season remains unrealized. The primary challenge lies in the inadequacy of conventional vision systems, such as standard RGB and RGB-D cameras, under the complex conditions of dense foliage. These systems struggle to function effectively amidst the dense leaf cover, failing to accurately identify necessary structural elements of the trees, making automation nearly impossible. In scenarios where trees are fully enveloped by canopy foliage, the vision systems cannot adequately penetrate the leaves or discern the crucial architectural elements required for tasks such as pruning and thinning. This severe limitation hinders not only the advancement but also the deployment of automated solutions during the peak of foliage growth. Consequently, there are currently no automated or robotic solutions that operate effectively during the green foliage canopy season for operation such as fruitlet thinning, underscoring a significant gap in orchard automation efforts. 

To address the challenge of limited visibility in orchard management during the canopy season, where many crucial tree parameters are obscured, the objective of this study is to perform 3D reconstruction and information fusion of tree parameters from dormant to canopy seasons. This is achieved by leveraging dormant season data to enhance the foundational vision capabilities during the dense foliage of the canopy season. This methodology employs a  vision based robotic platform (Figures \ref{fig:Figure1}e and \ref{fig:Figure1}f) which systematically collects high-resolution 3D images of apple trees during both the dormant and canopy seasons respectively. This platform was utilized across the two distinct seasons to gather essential structural data. 

\textbf{The specific objectives of this study are:}
\begin{itemize}
    \item To collect high-resolution dormant and canopy season data of apple trees from a commercial orchard using a vision based robotic system.
    \item To annotate and prepare a comprehensive image dataset with manual annotations of branches and trunks for training a deep learning model.
    \item To develop and validate a deep learning model for accurate detection and instance segmentation of tree trunks and branches in complex dormant season images.
    \item To reconstruct trees in 3D point clouds and create a detailed 3D map from the deep learning segmentation results.
    \item To register the 3D reconstructed model with canopy season images of the same trees using the Fast Generalized Iterative Closest Point (Fast GICP) algorithm.
    \item To analyze and validate the field-level performance of our approach by comparing it against ground truth measurements of branch diameter, trunk diameter, and branch spacing.
\end{itemize}

\section{Related Works}
Despite considerable progress in 3D segmentation and reconstruction technologies, particularly in analyzing tree structures during the dormant season, their application to canopy season imagery remains largely unexplored. This section reviews relevant works in three key areas to establish a foundation for the novel approach introduced in this study. The first subsection, \textbf{3D Segmentation and Reconstruction in Dormant Season}, examines existing research methodologies and outcomes focusing on the dormant phase of orchards. The second subsection, \textbf{Advancements in Image Matching Techniques for Agricultural Applications}, explores how current image matching technologies have been utilized in various agricultural settings and their potential to inform and enhance the methodological framework for canopy season analysis.

\subsection{3D Segmentation and Reconstruction in Dormant Season}
Traditional methods of modeling tree structures in the dormant season have predominantly relied on skeleton extraction techniques, pivotal for preserving the topology, centeredness, and computational efficiency of specialty crop trees \cite{bucksch2010skeltre}. The foundational model, treating trees as recursive branching structures, was proposed by Honda in 1971, setting the stage for further developments in this field \cite{honda1971description}. This concept was expanded by Boudon et al. in 2003, who developed tools for classifying geometric properties of tree branches, aiding in the creation of detailed bonsai tree graphics \cite{boudon2003interactive}. Subsequent enhancements by Runions et al. and Ganster and Klen extended these models to generate realistic 3D tree structures, enhancing user interaction through interactive interfaces \cite{runions2007modeling, ganster20081}. Despite these advancements, early digital models were primarily theoretical and did not address practical applications in real-world agricultural settings. In response to this gap, recent research has focused on incorporating real-world applications, employing advanced imaging techniques like stereo imaging to reconstruct critical phenotypic traits such as branch thickness and length, crucial for automated pruning \cite{karkee2015method, tabb2017robotic}. Additionally, techniques like structure from motion and scale invariant feature transform (SIFT) have been used to uncover hidden tree parameters, though they face challenges such as occlusions and variable lighting \cite{tan2007image, digumarti2018automatic}. The latest advances have focused on automatic tree reconstruction using skeletonization algorithms based on 3D point clouds, significantly enhancing the accuracy and applicability of these models in overcoming traditional method limitations \cite{ saha2016survey, akbar2016novel, zhang20163d}. 

Recent advancements in deep learning have propelled significant progress in the 3D segmentation and reconstruction of individual trees, even in complex natural environments. Liu et al. \cite{liu2021point} developed a novel trunk-growth (TG) method for point-cloud segmentation of trees in Shangri-La City's natural forests, using point normal vectors and Z-axis components to guide growth constraints. This method achieved an impressive average F-score of 0.96 for point‑cloud segmentation of individual trees in complex natural forest scenes, demonstrating its effectiveness across various tree types and complex forest scenes. However, such methods still face challenges in generalizing across different species and dense undergrowth conditions. Kok et al. \cite{kok2023obscured} introduced an obscured branch recovery framework combining Unet++ with the Point2Skeleton and a novel obscured branch recovery (OBR) algorithm for segmenting and reconstructing 3D tree branches obscured in natural orchard settings. Despite achieving significant segmentation accuracy, the geometric-based recovery still struggles with high error rates in complex scenes. Jiang et al. \cite{jiang2022tree} employed a thermal camera alongside a Faster R-CNN model to recognize tree trunks under various lighting conditions, achieving moderate success with an average detection error ranging from 0.16 m to 0.3 m at varying distances. This approach highlights the potential for using thermal imaging in low-light conditions, but also points to challenges in accuracy and camera orientation. These studies collectively push the boundaries of tree segmentation and reconstruction technology, offering robust frameworks for future research in automated forest management and agricultural robotics, albeit with noted limitations in accuracy, environmental adaptability, and computational demand.

Several recent studies have focused on deep learning-based segmentation methods for branch and trunk detection in fruit trees, aiming to automate labor-intensive agricultural tasks. Medeiros et al. \cite{medeiros2017modeling} developed a system using a laser sensor to model dormant fruit trees, employing a split-and-merge clustering algorithm that achieved 98\% accuracy in identifying primary branches, though it was limited by slow processing speeds. Majeed et al. \cite{majeed2018apple} introduced a semantic segmentation approach using Kinect V2 and SegNet to segment trunks and branches in apple trees for trellis training, reaching high accuracies (0.92 for trunks and 0.93 for branches) but facing challenges with IoU scores for complex branches. Tong et al. \cite{tong2023image} proposed a pruning point localization system based on SOLOv2 instance segmentation, achieving an accuracy of 87.2\% in identifying pruning points with low error rates, although manual pruning rule adaptation limited scalability. Zhang et al. \cite{zhang2021computer} implemented a computer vision system with Deeplab v3+ ResNet-18 to detect shaking points for automated apple harvesting, with ResNet-18 achieving strong IoU and boundary F1-scores for trunk and branch identification. Lastly, Wan et al. \cite{wan2022real} developed a real-time branch detection and reconstruction mechanism using Branch-CNN, achieving high precision and F1-scores and an impressive 22.7 FPS processing speed, but its practical application requires further validation across diverse crop environments. Collectively, these studies highlight the potential of deep learning for efficient segmentation and reconstruction in agricultural automation, while also revealing common limitations in environmental adaptability, segmentation accuracy, and real-time processing feasibility.

\subsection{Advancements in Image Registration Techniques in 3D}
The Iterative Closest Point (ICP) algorithm, first introduced by Besl and McKay \cite{besl1992method}, revolutionized 3D shape registration by providing an efficient and robust approach for aligning point clouds and surfaces. Their method, which minimized mean-square distances between corresponding points, quickly became foundational in 3D modeling, computer vision, and robotics, especially for tasks requiring alignment across six degrees of freedom. Despite its effectiveness, the original ICP algorithm faced challenges in terms of speed and sensitivity to initial pose estimates. Subsequent research focused on improving these aspects, leading to a variety of ICP variants. Rusinkiewicz and Levoy \cite{rusinkiewicz2001efficient} explored and categorized these ICP variants, proposing a combination optimized for high-speed alignment in real-time applications, which expanded the algorithm’s usability in dynamic environments. Ezra et al. \cite{ezra2008performance} provided rigorous performance analysis, demonstrating that ICP’s iterative convergence under root mean square (RMS) and Hausdorff distances could be bounded, although the algorithm’s convergence rate varies with point set characteristics. Gelfand et al. \cite{gelfand2003geometrically} addressed the issue of unstable convergence by proposing a geometrically stable sampling method, which improved alignment accuracy in featureless or noisy datasets by minimizing uncertainties in pose estimation. Jost et al. \cite{jost2002fast} further improved ICP’s efficiency by implementing a k-D tree search to optimize point matching, reducing time complexity from O(N²) to O(N log N). Collectively, these advancements established a robust foundation for modern ICP applications, enhancing speed, accuracy, and stability in 3D shape registration.

Recent studies have applied ICP algorithms across a range of agricultural applications, enhancing the accuracy and functionality of point cloud registration in plant and tree monitoring tasks. Zhang et al. \cite{zhang2022point} developed a conical surface fitting ICP method to reconstruct 3D models of maize plants, achieving a registration error of 1.98 mm, essential for reliable growth observations and machinery research. Yuan et al. \cite{yuan2022gnss} introduced a GNSS-IMU-assisted ICP method for UAV-LiDAR registration in peach trees, improving tree parameter estimation accuracy by 67\% compared to traditional methods, demonstrating a low-cost alternative for crop inspection. Zhou et al. \cite{zhou2020point} used calibration balls to improve registration accuracy in crops, achieving fitness scores of approximately 0.0001 for trunk and branch alignment, which aids in ensuring precision in multi-view scans of agricultural objects. Jiang et al. \cite{jiang2024navigation} developed an NDT ICP-based navigation system for orchard spraying robots, achieving sub-meter level navigation accuracy, proving effective in autonomous orchard applications. Schut et al. \cite{schut2024joint} applied a joint 2D-3D registration method to align CT and photographic slices of apples, with a mean error of 1.47 mm, beneficial for assessing internal fruit disorders. Cui et al. \cite{cui2023situ} utilized ICP-based techniques in elemental imaging of plant leaves, enhancing quantitative accuracy in in-situ analysis of element distribution. Zhang et al. \cite{zhang2023spatio} developed a registration model to trace plant growth, capturing non-rigid deformations effectively for monitoring plant structure. Lastly, Zhang et al. \cite{zhang2024image} integrated image segmentation with Fast Global Registration to detect obscured branches in orchards, achieving a 24\% increase in corresponding points, useful for robotics in natural orchards. Together, these studies illustrate ICP's versatility in enhancing agricultural data accuracy and efficiency.

\section{Methodology}
\begin{figure*}[h!]
    \centering
    \includegraphics[width=7 in, height=4 in]{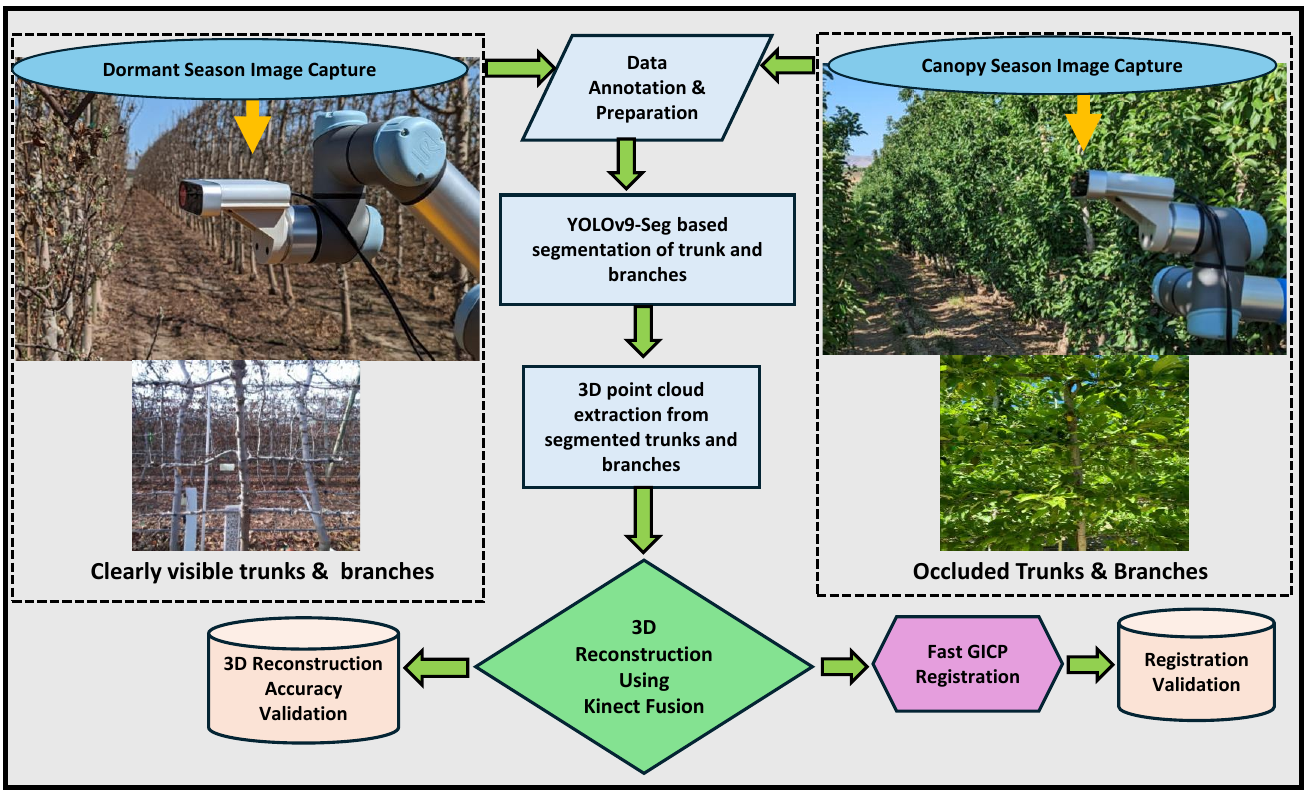}
    \caption{Workflow diagram illustrating the methodology, from data acquisition and YOLOv9 segmentation through 3D reconstruction with Kinect Fusion, to alignment with Fast GICP for seasonal model integration.}
    \label{fig:Figure2}
\end{figure*}

This study employed a methodology focused on information fusion to enhance structural analysis and management of orchards across the canopy season using dormant season information. High-resolution images along with ground truth information about the trees were captured using a robotic platform equipped with an Azure Kinect DK sensor, as detailed in Figure \ref{fig:Figure1}e for the dormant season and Figure \ref{fig:Figure1}f for the canopy season. During the dormant season, the images provided clear views of tree trunks and branches, while in the canopy season, only partial views were possible beneath dense foliage. As illustrated in Figure \ref{fig:Figure2}, these datasets were then annotated and utilized to train the YOLOv9 deep learning model for precise multi-class segmentation. The segmented data facilitated the generation of 3D point clouds by extracting depth data, which were then reconstructed into detailed tree skeletons using Kinect Fusion. This approach of integrating seasonal image and structural data allowed for detailed comparison and alignment of tree structures across seasons. The Fast Generalized Iterative Closest Point (FGICP) technique was applied to achieve precise alignment of the trees from the two different seasons, creating an integrated view in 3D environment. The methodology overview leads into specific subsections that detail each phase of the study that led to effective information fusion and 3D reconstruction of apple tree models: \textit{Image and Data Acquisition}, \textit{Image Annotation, Preprocessing, and Preparation}, \textit{YOLOv9 Deep Learning Model Training}, \textit{3D Reconstruction Using KinectFusion}, \textit{Validation and Verification}, and \textit{Alignment \& Registration Using Fast GICP (FGICP)}.

\subsection{Study Site and Data Acquisition}
This study was conducted in a commercial apple orchard owned and operated by Allan Brothers Fruit Company, located in Prosser, Washington State, USA. Planted in 2009, the orchard features Scilate apple cultivars arranged with a row spacing of 9.0 feet and a plant spacing of 3.0 feet and was trained in a V-trellis architecture. A vision-based robotic platform was utilized (as shown in Figures \ref{fig:Figure1}e and \ref{fig:Figure1}f), comprising a Microsoft Azure Kinect DK sensor (as shown in Figure \ref{fig:Figure2} (https://azure.microsoft.com/en-us/products/kinect-dk)), mounted on a UR5e industrial robotic arm (Universal Robotics) based in the USA. The UR5e arm was retrofitted on a Warthog ground robot (Clearpath Robotics, Ontario, Canada), enabling precise maneuverability and stability during image collection across the orchard during the two seasons.  It is a compact and advanced sensing device with a range of critical features for 3D data acquisition. Measuring 126 x 103 x 39 mm and weighing 440g, it combines a 1-megapixel depth camera and a 12-megapixel RGB camera, supported by a 7-microphone circular array for audio capture. The depth camera employs Time-of-Flight (ToF) technology to precisely measure distances by calculating light travel time, enhancing the depth map’s accuracy essential for reconstructing complex tree structures. The RGB camera includes a CMOS sensor with a rolling shutter, providing high-resolution color imaging at 12MP, suitable for detailed feature extraction on tree trunks and branches. Using this machine vision camera, the images were collected across two the two distinct seasons. The dormant season images (as shown earlier in Figure \ref{fig:Figure1}e) were collected in the month of December 2023, while the canopy season images were collected in the month of June 2024 (as illustrated earlier in Figure \ref{fig:Figure1}f). Initially, a total of 859 images were collected from various trees in the commercial orchard. The robotic platform maintained a consistent distance of 2.5 feet from the canopy, navigating such that the camera's view remained parallel to the trees.

\subsection{Image Annotation, Preprocessing, and Preparation}
\begin{figure*}[h!]
    \centering
    \includegraphics[width= 7 in, height= 3 in]{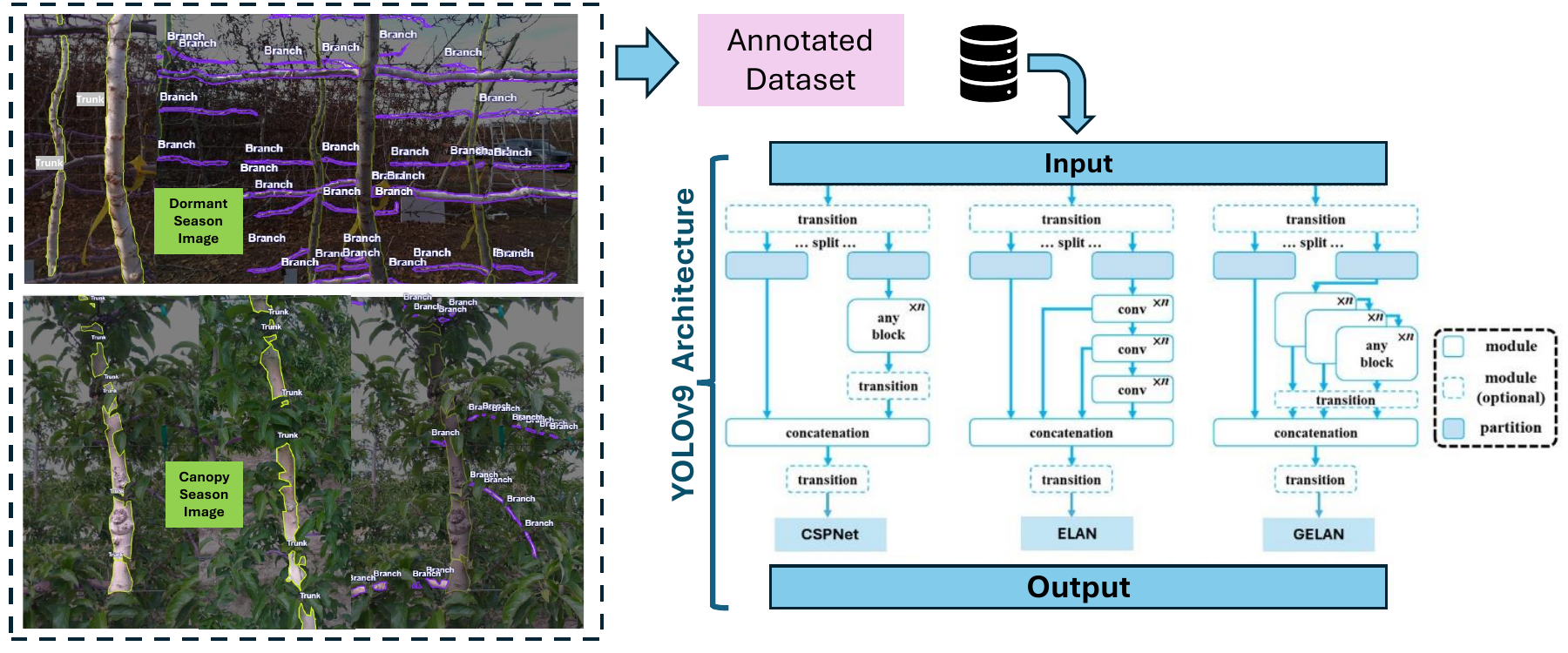}
    \caption{Illustrating data preparation for training deep learning to detect and segment tree trunks and branches in the orchard images in two distinct seasons}
    \label{fig:MethodsFigure2}
\end{figure*}

To prepare a robust dataset for deep learning segmentation, a total of 859 images were carefully annotated to label tree trunks and branches using roboflow (Roboflow, Des Moines, IOWA, USA). Figure \ref{fig:MethodsFigure2}a illustrates representative images from both the dormant and canopy seasons, with the left side showing the unobstructed trunk and branch visibility typical of dormant season images and the right side showing the partial occlusion caused by foliage in the canopy season. This labor-intensive annotation process required 172 hours of manual effort, carefully delineating each tree structure to provide a high-quality dataset for training the YOLOv9-seg model.

The annotation process was conducted using Roboflow software, which allowed for detailed labeling and export in the YOLOv9 segmentation format, supporting efficient model training and testing. Figure \ref{fig:MethodsFigure2}b provides a view of the manual annotation overlay, where visible trunk and branch structures in both dormant and canopy seasons were labeled. For canopy season images, only visible parts of the trunks and main branches within each camera frame were annotated, representing the limited visibility due to foliage. In contrast, dormant season images captured clear views of the entire trunk and primary branches, enabling comprehensive labeling for these images.

The dataset was then divided into training and validation subsets, following an 80:20 split. Although 859 images were collected, the training set comprised 687 images, including 553 from the canopy season where trees were largely obscured by leaves, and 134 from the dormant season which provided clear views of the tree structure. The remaining images were excluded as they lacked adequate trunk and branch details for annotations or were repetitive, similar to previous frames, ensuring the cleanliness of the data. The validation set included 142 canopy season images and 20 dormant season images, ensuring a diverse seasonal representation for robust testing.

Each tree structure was outlined with precision, employing polygon-based annotations for both trunk and branch elements. These detailed annotations were exported in YOLOv9 data format for further training the model. 

\subsection{YOLOv9 Deep Learning Model Training and Performance Evaluation}
YOLOv9 \cite{wang2024yolov9} builds upon these advancements and provides a balanced approach suitable for our application as the introduction of the Programmable Gradient Information (PGI) mechanism addresses key challenges in deep learning by managing gradient flow for enhanced feature retention across layers. YOLOv9 also employs the innovative Generalized Efficient Layer Aggregation Network (GELAN) as illustrated in the architecture diagram in Figure \ref{fig:MethodsFigure2}, a lightweight architecture designed to minimize information loss during deep network processing. The model was trained on a workstation with an Intel Xeon® W-2155 CPU @ 3.30 GHz x20 processor, NVIDIA TITAN Xp Collector's Edition/PCIe/SSE2 graphics card, 31.1 GiB memory, and Ubuntu 16.04 LTS 64-bit operating system. Three different configurations of YOLOv9 were trained : YOLOv9-seg-dseg, YOLOv9-gelan-c-seg, and YOLOv9-gelan-c-dseg. Summary of hyperparameters used in training these models are presented in Table \ref{tab:hyperparameters} as: 

\begin{table}[h!]
    \centering
    \caption{Summary of YOLOv9 model training hyperparameters, including optimization settings, loss function gains, augmentation parameters, and probability factors, optimized for accurate segmentation of tree trunk and branches in varied orchard conditions.}
    \begin{tabular}{|p{2.5cm}|p{1cm}|p{3cm}|p{1cm}|}
        \hline
        \textbf{Hyperparameter} & \textbf{Value} & \textbf{Hyperparameter} & \textbf{Value} \\
        \hline
        Initial Learning Rate ($lr0$) & 0.01 & Final Learning Rate Factor ($lrf$) & 0.01 \\
        Momentum & 0.937 & Weight Decay & 0.0005 \\
        Warmup Epochs & 3.0 & Warmup Momentum & 0.8 \\
        Warmup Bias LR & 0.1 & Box Loss Gain & 7.5 \\
        Class Loss Gain ($cls$) & 0.5 & Positive Weight ($cls\_pw$) & 1.0 \\
        Object Loss Gain ($obj$) & 0.7 & Obj Positive Weight ($obj\_pw$) & 1.0 \\
        DFL Loss Gain ($dfl$) & 1.5 & IoU Threshold ($iou\_t$) & 0.20 \\
        Anchor Threshold & 5.0 & Focal Loss Gamma & 0.0 \\
        HSV Hue ($hsv\_h$) & 0.015 & HSV Saturation ($hsv\_s$) & 0.7 \\
        HSV Value ($hsv\_v$) & 0.4 & Rotation ($degrees$) & 0.1 \\
        Translation & 0.1 & Scale & 0.5 \\
        Shear & 0 & Perspective & 0.0 \\
        Flip Up-Down ($flipud$) & 0.5 & Flip Left-Right ($fliplr$) & 0.5 \\
        Mosaic & 1.0 & Mixup & 0 \\
        \hline
    \end{tabular}
    \label{tab:hyperparameters}
\end{table}

To assess the performance of YOLOv9 for instance segmentation of tree trunks and branches in both dormant and canopy seasons, a set of standard evaluation metrics was applied. These metrics provide quantitative insights into the segmentation accuracy, overlap quality, and balance between correctly and incorrectly identified regions. The primary metric used was the Mean Intersection over Union (MIoU), complemented by Precision, Recall, and F1-score calculations to evaluate the segmentation quality comprehensively.

The MIoU metric, which measures the overlap between predicted and ground truth segmentation areas, is defined as follows:
\begin{equation}
    \text{MIoU} = \frac{\text{Area}_{\text{Overlap}}}{\text{Area}_{\text{Union}}} = \frac{\text{TP}}{\text{FP} + \text{TP} + \text{FN}}
\end{equation}
where TP, FP, and FN represent the True Positives, False Positives, and False Negatives, respectively. Here, the area of overlap refers to the intersection between the predicted and actual trunk and branch regions, while the area of union accounts for the total area covered by both.

In addition to MIoU, Precision, Recall, and F1-score were also calculated. Precision indicates the proportion of correctly identified trunk and branch pixels out of all pixels predicted as trunk or branch:
\begin{equation}
    \text{Precision} = \frac{\text{TP}}{\text{TP} + \text{FP}}
\end{equation}

Recall represents the fraction of actual trunk and branch pixels that were correctly predicted:
\begin{equation}
    \text{Recall} = \frac{\text{TP}}{\text{TP} + \text{FN}}
\end{equation}

The F1-score, which provides a balance between Precision and Recall, is computed as:
\begin{equation}
    \text{F1-score} = 2 \times \frac{\text{Precision} \times \text{Recall}}{\text{Precision} + \text{Recall}}
\end{equation}
\subsection{3D Reconstruction Using KinectFusion}

The KinectFusion algorithm was applied to achieve real-time 3D reconstruction of apple trees by processing segmented data produced by the YOLOv9 model. During data collection, depth information was captured using a Kinect sensor in both dormant and canopy seasons. For the dormant season, clear visibility allowed for the segmentation of all trunks and branches, while in the canopy season, only the trunks were partially visible due to foliage obstruction. These segmented depth images were transformed into 3D point clouds, denoted as \( P \), where each point had spatial coordinates \( (x, y, z) \). The KinectFusion system utilized these points to incrementally build a comprehensive 3D model, continuously integrating new data into a global volumetric model \( V \) as described by:
\begin{equation}
    V_{\text{new}} = \text{KinectFusion}(V, P)
    \label{eq:fusion}
\end{equation}

Each dataset, representing seasonal variations, was reconstructed separately to create distinct 3D models for the dormant and canopy seasons. This approach enabled precise validation of field-measured values, such as trunk diameters, branch diameters, and branch spacing, against the ground truth measurements collected in the orchard.

Following the initial YOLOv9 segmentation and 3D reconstruction via KinectFusion, the accuracy of the reconstructed models was verified. The reconstruction accuracy was validated by comparing KinectFusion-derived measurements with ground truth data for three specific metrics: branch diameter, trunk diameter, and branch spacing. These measurements were collected in-field using digital calipers and measuring tapes. The YOLOv9 and KinectFusion-predicted values were then extracted from the point clouds using CloudCompare, a specialized 3D measurement software. The root mean squared error (RMSE) and mean absolute error (MAE) were calculated to assess discrepancies between predicted values and actual measurements, with both metrics offering insight into the reconstruction accuracy.

Additionally, KinectFusion refined the quality of the 3D model by adjusting for occlusions and movement variations in the orchard environment. GPU-based techniques were employed to process the data in real-time, maintaining high fidelity in the 3D structure, while RGB data from the Kinect sensor enhanced texture mapping for realistic visual representation. This robust 3D reconstruction provided an accurate model essential for detailed analysis and assessment of orchard tree structure and health.

\subsection{Validations and Verification}
In this study,in-field measurements of  tree trunk diameters, branch diameters, and branch spacing, were thoroughly validated using 3D point clouds generated by the YOLOv9 segmentation model and Kinect Fusion reconstruction process. CloudCompare software was employed to extract and analyze these metric values directly from the reconstructed 3D models. This analysis involved a systematic comparison of predicted values from the 3D model with ground truth measurements collected in the orchard. Precise digital calipers and measuring tapes were used during the data acquisition phase to ensure accuracy.

\begin{figure*}[h!]
    \centering
    \includegraphics[width=7 in, height=3 in]{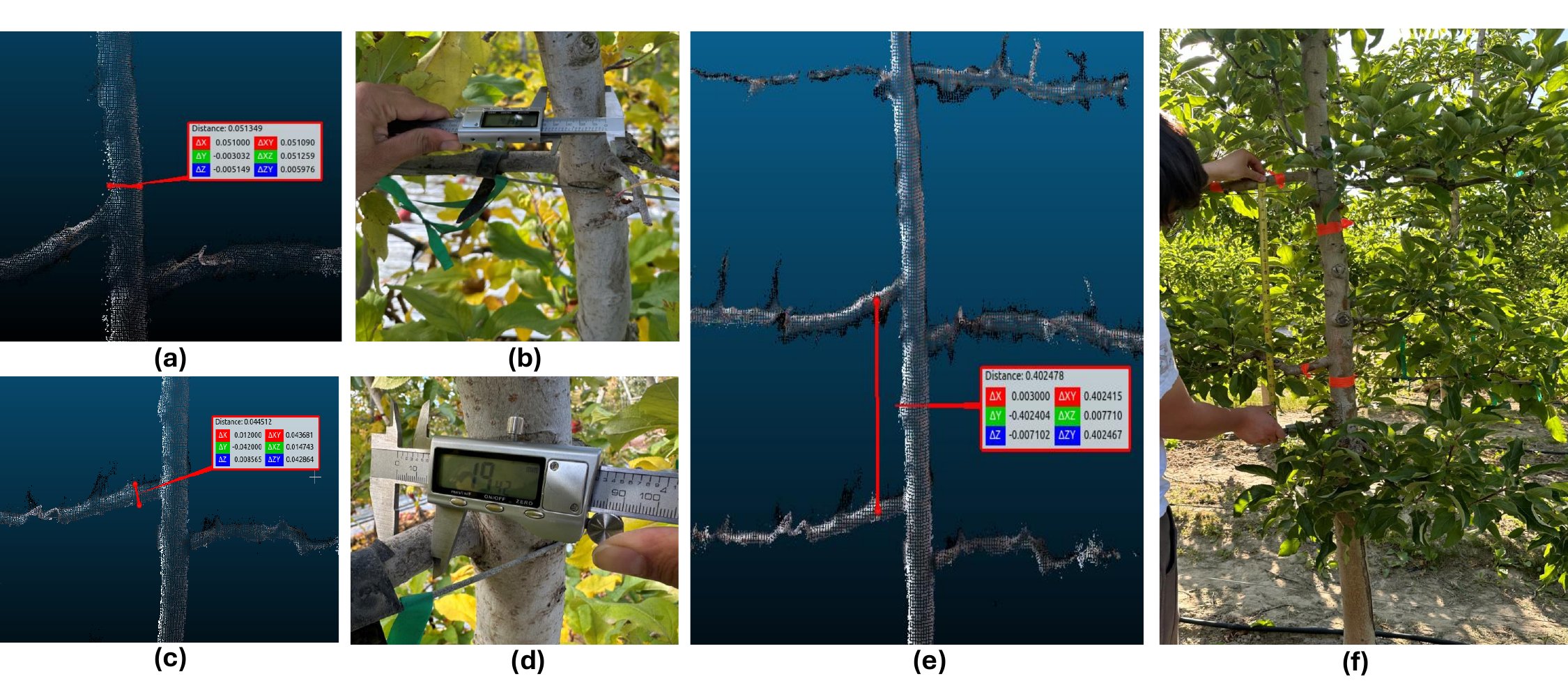}
    \caption{Validation and Verification of 3D Reconstructions with Ground Truth Measurements: (a) Illustrates the measurement of trunk diameter within CloudCompare software, showing the accuracy of post-processed YOLOv9 segmentation and KinectFusion 3D reconstruction against the actual tree structure; (b) Depicts the ground truth collection of trunk diameter using a vernier caliper in the field during the canopy season, providing a direct comparison to validate the 3D model; (c) Displays the measurement of branch diameter within CloudCompare, verifying the dimensional accuracy of the reconstructed branches; (d) Shows the field measurement of branch diameter with a vernier caliper, used to confirm the 3D model's accuracy; (e) Presents the measurement of branch spacing in CloudCompare after YOLOv9 segmentation and KinectFusion 3D reconstruction, highlighting the detailed capture of spatial relationships; (f) Illustrates ground truth validation of branch spacing measured with a measuring tape in the field, two inches away from the trunk, ensuring comprehensive verification of the 3D reconstructed model.}
    \label{fig:Methodvalid}
\end{figure*}

Figure \ref{fig:Methodvalid}a illustrates the measurement of trunk diameter within the CloudCompare software, following post-processing from the YOLOv9 segmentation and KinectFusion 3D reconstruction. Correspondingly, Figure \ref{fig:Methodvalid}b depicts the ground truth collection of trunk diameter using a vernier caliper in the field during the canopy season. This ensures a direct comparison between the reconstructed model and actual physical dimensions.

Similarly, Figure \ref{fig:Methodvalid}c displays the measurement of branch diameter within CloudCompare, offering a visual validation against the field measurements shown in Figure \ref{fig:Methodvalid}d, where branch diameters were measured using a vernier caliper during the canopy season. These comparisons are crucial for verifying the accuracy of the 3D reconstructions against real-world measurements.

Furthermore, Figure \ref{fig:Methodvalid}e shows the measurement of branch spacing using the CloudCompare software post-YOLOv9 segmentation and KinectFusion 3D reconstruction. Figure \ref{fig:Methodvalid}f illustrates the ground truth validation, where the actual spacing between branches, situated two inches from the trunk, was measured in the field with a measuring tape. This juxtaposition of 3D model data with field measurements substantiates the model’s reliability in replicating true orchard tree geometries, highlighting the effectiveness of the employed methodologies in capturing precise structural details of the trees.
The validation metrics included both root mean squared error (RMSE) and mean absolute error (MAE), calculated to assess alignment between the reconstructed and actual measurements, thus indicating the reliability of the 3D model. The RMSE, as defined in Equation \ref{eq:rmse}, emphasizes larger discrepancies, while MAE, shown in Equation \ref{eq:mae}, provides an average absolute error for a balanced measure of prediction accuracy:
\begin{equation}
    \text{RMSE} = \sqrt{\frac{1}{N} \sum_{i=1}^N (y_i - \hat{y}_i)^2}
    \label{eq:rmse}
\end{equation}
\begin{equation}
    \text{MAE} = \frac{1}{N} \sum_{i=1}^N |y_i - \hat{y}_i|
    \label{eq:mae}
\end{equation}

In these equations:
- \( y_i \) represents each ground truth measurement, specifically trunk diameter, branch diameter, or branch spacing obtained through direct in-field measurement.
- \( \hat{y}_i \) denotes the corresponding predicted value measured from the 3D model.
- \( N \) is the total number of measurement samples.

The RMSE and MAE calculations confirmed the high fidelity and accuracy of the 3D reconstructions, showing close alignment with physical measurements. This consistency and precision validate the efficacy of the YOLOv9 segmentation and Kinect Fusion methodology, demonstrating its potential for broader applications in precision agriculture. By confirming the model’s accuracy in replicating actual field conditions, this validation process reinforces confidence in advanced technologies for detailed orchard management and supports further innovation in agricultural practices.

\subsection{Alignment and Registration Using Fast GICP (FGICP)}
Upon the completion of the 3D model using KinectFusion, the Fast Generalized Iterative Closest Point (FGICP) algorithm was utilized to align the dormant season models with the canopy season images. FGICP enhances the traditional GICP by incorporating fast voxelization, reducing computational complexity and enabling real-time performance crucial for navigating dynamic orchard environments. The FGICP operates by voxelizing the point cloud into a structured grid, where each voxel represents an aggregated distribution of the points it contains. The alignment process iteratively refines the transformation between the reconstructed model and the target model, minimizing the discrepancies between corresponding voxels:
\begin{equation}
    T = \text{argmin}_{T} \sum_{i} \left( d_i^T \left( C_{B_i} + T C_{A_i} T^T \right)^{-1} d_i \right)
    \label{eq:GICP}
\end{equation}

Here, \( d_i \) is the displacement vector for the \( i \)-th voxel, \( C_{A_i} \) and \( C_{B_i} \) are the covariance matrices of the corresponding voxels in the source and target models, and \( T \) is the transformation matrix being optimized.

\subsection{Fast GICP Registration Validation}
In this stage, the alignment accuracy of the 3D tree models, reconstructed from the KinectFusion outputs, was validated by calculating the Fast GICP fitness score for each tree across dormant and canopy seasons. The fitness score, an indicator of registration precision, was calculated as follows:

\begin{equation}
    \text{Fitness Score} = \frac{1}{N} \sum_{i=1}^{N} \left\| p_i - T(q_i) \right\|
    \label{eq:fitness_score}
\end{equation}

where \( N \) represents the total number of point pairs used in alignment, \( p_i \) denotes a point in the target model, and \( q_i \) represents a corresponding point in the source model. The transformation matrix \( T \), optimized through Fast GICP, was applied to \( q_i \) to align it with \( p_i \). The fitness score provides a mean displacement value between aligned points, where lower values signify more precise registration. This validation metric confirmed the effectiveness of the FGICP alignment, essential for accurate comparisons between seasonal tree structures.

\section{Results and Discussion}
\subsection{YOLOv9 Segmentation Results for Tree Trunk and Branches}
\begin{figure*}[h!]
    \centering
    \includegraphics[width=7 in, height=4 in]{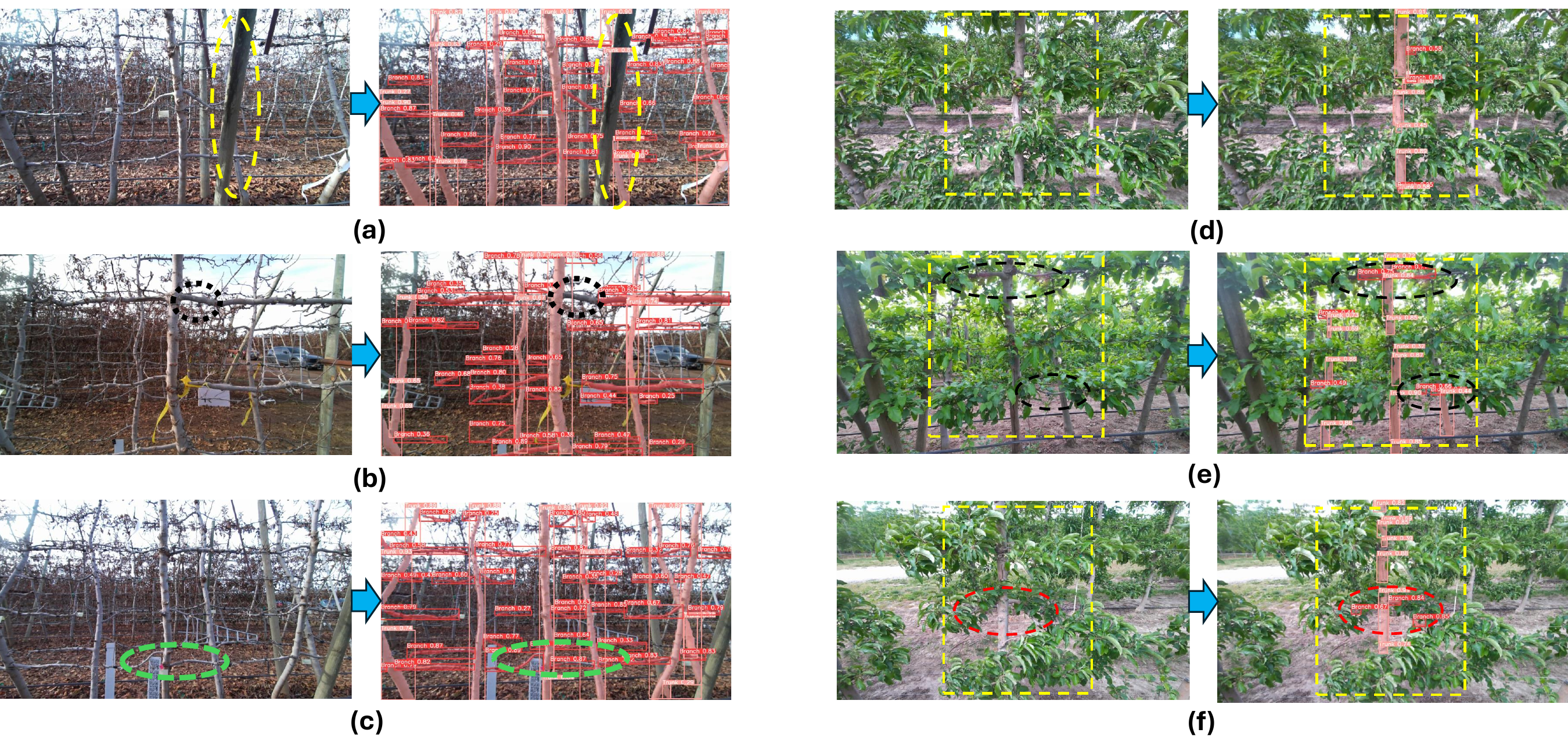}
    \caption{Visualization of segmentation results of YOLOv9 model in dormant and canopy seasons: (a) Shows the effective segmentation of tree trunks and branches during the dormant season, with a comparison between the original image (left) and segmented outcome (right). (b) Illustrates some limitations in the YOLOv9-gelan-c-seg variant during dormant season segmentation, with missed branch foreground highlighted in black. (c) Displays partial segmentation inconsistencies within a green dotted circle by the YOLOv9-c-dseg variant. (d) Depicts only visible trunk segmentation by the YOLOv9-gelan-c-dseg variant during the canopy season. (e) Minimal branch detection in canopy season by the YOLOv9-gelan-c-seg variant within black dotted circles. (f) Shows sporadic branch detection in densely foliated conditions by the YOLOv9-c-dseg variant, highlighted within red circles.}
    \label{fig:Figure1Result}
\end{figure*}

The effectiveness of the YOLOv9 model in segmenting dormant season trees is as illustrated in Figures \ref{fig:Figure1Result}a, b and c where the segmentation results for dormant branches were particularly promising, with most of the trunk and branches being accurately delineated. This precise segmentation underscores the model’s capability to discern and outline even the subtle distinctions between the tree components during the dormant season. Conversely, as shown in Figures \ref{fig:Figure1Result}d, e and f, the YOLOv9 model's performance during the canopy season was notably constrained. Segmentation was effectively limited to parts of the trunk that remained visible; these sections represent the elements typically detectable by the human eye under normal viewing conditions. This differential outcome highlights the challenges posed by dense foliage in accurately segmenting tree structures and points to the need for tailored approaches that adapt to varying seasonal visibility.

In the detailed analysis of the segmentation results captured during the dormant season, Figure \ref{fig:Figure1Result}a demonstrates the exceptional capability of the YOLOv9-gelan-c-dseg configuration. The left side of the subfigure presents the original image captured in December, with the right side displaying the segmented outcome. Notably, this model adeptly identifies and segments both branches and trunks, despite the presence of occluding objects such as training poles used in commercial Scilate orchard apple cultivation. The model's robustness is particularly evident as it accurately continues the segmentation of tree trunks through the yellow circled regions where poles obscure the background. This impressive performance underscores the YOLOv9-gelan-c-dseg variant's effectiveness in handling complex scenes.

Subsequently, Figure \ref{fig:Figure1Result}b, showcasing the YOLOv9-gelan-c-seg variant, reveals some limitations in segmentation accuracy. The original image on the left and the deep learning processed image on the right highlight areas, encircled in black, where the model fails to segment a prominent branch foreground. This suggests a potential improvement area for this variant, where increased training samples could enhance detection sensitivity, particularly for elements in the visual forefront.

Lastly, the outcomes from the YOLOv9-c-dseg variant illustrated in Figure \ref{fig:Figure1Result}c depict a competent but partially inconsistent segmentation performance within the green dotted circle. While the model successfully segments a significant portion of trunks and branches, it exhibits partial segmentation in areas crucial for precise Limb Cross Section Area (LCSA) determinations, a key metric for automated orchard operations like fruitlet thinning. Addressing this through augmented training datasets could refine the model’s accuracy, ensuring more reliable and comprehensive segmentation necessary for advanced agricultural practices.

In the canopy season, the segmentation challenges are magnified due to dense foliage, as evidenced by the results presented in right side of Figure \ref{fig:Figure1Result}. Figure \ref{fig:Figure1Result}d illustrates the YOLOv9-gelan-c-dseg variant's performance, where only visible trunks could be reliably segmented. The original image and its segmented counterpart within the yellow rectangular dotted region highlight the model's limited capacity to detect branches under heavy leaf coverage. No significant branch information was detected, indicating the variant's specificity to trunk segmentation under these conditions.

Figure \ref{fig:Figure1Result}e, processed by the YOLOv9-gelan-c-seg variant, shows minimal branch detection within black dotted circles. The identified branch segments are too insubstantial for practical applications in crop load management, underlining the need for enhanced sensitivity in branch detection algorithms. This suggests that current training data may not sufficiently represent the complexity of fully leafed scenes, requiring augmentation to improve utility.

Similarly, Figure \ref{fig:Figure1Result}f, depicting results from the YOLOv9-c-dseg variant, demonstrates a sporadic detection of branches within the red circled region. Although the segmented branches are visible, they appear as disjointed points with no practical segmentation value, reflecting the overarching challenge of obtaining useful branch data in densely foliated conditions.

Despite these limitations, all three YOLOv9 model variants performed commendably in terms of trunk detection, as visible trunks were consistently identified and segmented across all subfigures. This capability underscores the models' effectiveness in scenarios where only limited portions of the tree structure are visible, akin to what is typically perceivable by the naked human eye during peak foliage. The findings underscore the necessity for further model training and refinement to better handle the complex visual environments presented by orchard canopies during the growing season.

\subsubsection{Performance Metrics Evaluation}
The YOLOv9 segmentation model was evaluated across three variants to determine their effectiveness in segmenting tree structures into trunk and branch categories. The mask precision, which is critical for assessing the accuracy of the segmentation boundaries, revealed differential performance across these variants. The YOLOv9-seg-dseg variant achieved mask precision values of 0.67, 0.58, and 0.77 for all, branch, and trunk categories, respectively. In comparison, the YOLOv9-gelan-c-seg variant recorded slightly lower mask precision values of 0.65, 0.56, and 0.74 for the same categories. The last variant, YOLOv9-gelan-c-dseg, showed mask precision values of 0.66, 0.57, and 0.76. Among these, the YOLOv9-seg-dseg variant demonstrated the best overall performance, particularly notable in the critical trunk category with a precision of 0.77. This suggests that the YOLOv9-seg-dseg model is the most effective for accurate segmentation of tree structures in orchard environments, making it a preferable choice for detailed analysis and monitoring tasks.

\begin{figure*}[h!]
    \centering
    \includegraphics[width=6 in, height=4 in]{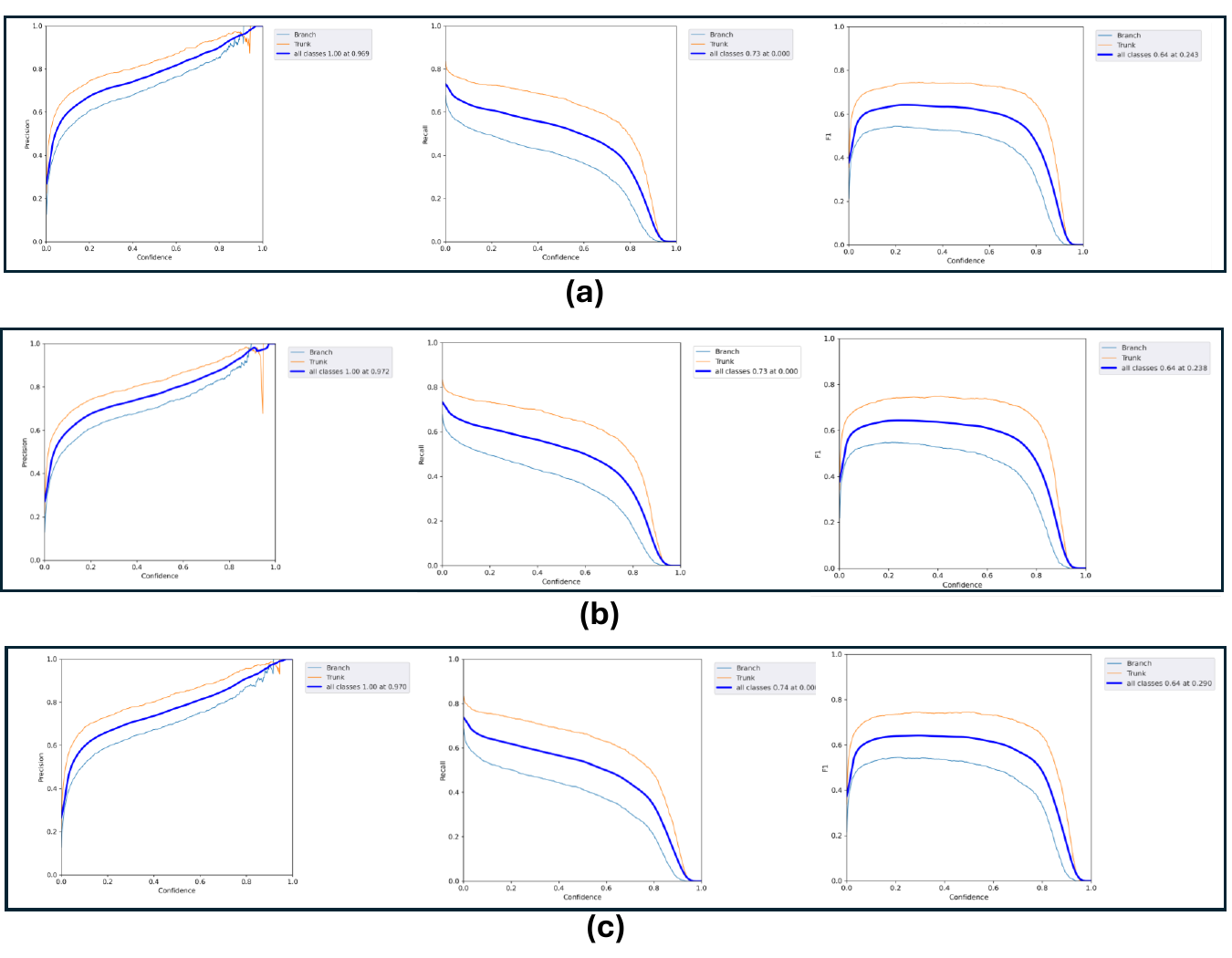}
    \caption{Recall Metrics Curve for all three variants of YOLOv9 segmentation for tree trunk and branches: a) YOLOv9-c-dseg ; b)YOLOv9-c-seg and c) YOLOv9 Gelan-c-dseg }
    \label{fig:Figure3Result}
\end{figure*}

Likewise, the effectiveness of the YOLOv9 model variants was also evaluated based on their recall values as illustrated in Figure \ref{fig:Figure3Result}, which measure the ability to correctly identify relevant instances across all, branch, and trunk categories. The YOLOv9-seg-dseg variant demonstrated recall scores of 0.56, 0.42, and 0.70 respectively for these categories. Meanwhile, the YOLOv9-gelan-c-seg variant showed slightly improved recall values of 0.57 for all categories, 0.43 for branches, and 0.71 for trunks, indicating a marginal enhancement over the seg-dseg variant. The gelan-c-dseg variant matched the gelan-c-seg in recall for all and branches categories but significantly dropped in trunk category to 0.52. Notably, both gelan-c-seg and gelan-c-dseg variants achieved the highest recall for the trunk category at 0.71, suggesting their slightly better performance in identifying this crucial category under various conditions. However, the drop in trunk recall for the gelan-c-dseg variant indicates a potential trade-off between precision and recall in this model. These insights reveal that the YOLOv9-gelan-c-seg variant stands out as the best performer in terms of recall, particularly for the trunk category, which is essential for effective orchard management applications. The detailed reports on mask and box metrics for the detection and segmentation of trunks and branches are presented in Table \ref{tab:performance}.

\begin{table*}[h]
\centering
\caption{YOLOv9 Variants Performance for Precision and Recall by Category}
\label{tab:performance}
\resizebox{6in}{!}{ % Resizes the table to 6 inches wide
\begin{tabular}{|c|c|c|c|c|c|c|c|c|c|c|c|c|}
\hline
\textbf{Model Variant} & \multicolumn{3}{c|}{\textbf{Mask Precision}} & \multicolumn{3}{c|}{\textbf{Box Precision}} & \multicolumn{3}{c|}{\textbf{Mask Recall}} & \multicolumn{3}{c|}{\textbf{Box Recall}} \\ \cline{2-13} 
 & \textbf{All} & \textbf{Branch} & \textbf{Trunk} & \textbf{All} & \textbf{Branch} & \textbf{Trunk} & \textbf{All} & \textbf{Branch} & \textbf{Trunk} & \textbf{All} & \textbf{Branch} & \textbf{Trunk} \\ \hline
YOLOv9-seg-dseg & 0.67 & 0.58 & 0.77 & 0.75 & 0.70 & 0.79 & 0.56 & 0.42 & 0.70 & 0.64 & 0.53 & 0.75 \\ \hline
YOLOv9-gelan-c-seg & 0.65 & 0.56 & 0.74 & 0.74 & 0.70 & 0.79 & 0.57 & 0.43 & 0.71 & 0.65 & 0.54 & 0.75 \\ \hline
YOLOv9-gelan-c-dseg & 0.66 & 0.57 & 0.76 & 0.75 & 0.70 & 0.80 & 0.57 & 0.43 & 0.52 & 0.63 & 0.52 & 0.52 \\ \hline
\end{tabular}
}
\end{table*}

Likewise, interms of the mean Average Precision (mAP@50) metric (Table \ref{tab:map50}), which was evaluated for both mask and box predictions across different categories (All, Branch, and Trunk), the YOLOv9-seg-dseg variant demonstrated robust performance, especially for trunk detection with mAP@50 scores reaching 0.75 for mask and 0.78 for box, indicating a high level of precision in critical structural detection. The YOLOv9-gelan-c-seg and YOLOv9-gelan-c-dseg variants showed slightly lower performance in branch detection with mAP@50 scores of 0.418 and 0.407 for mask, respectively.  However, all variants maintained consistent scores for trunk detection in box metrics, suggesting stable performance across models. Detailed report on mAP@50 metrics are presented in Table \ref{tab:map50}. 

\begin{table*}[h]
\centering
\caption{mAP@50 Values for YOLOv9 Variants by Category}
\label{tab:map50}
\resizebox{6in}{!}{
\begin{tabular}{|c|c|c|c|c|}
\hline
\textbf{Metric} & \textbf{Category} & \textbf{YOLOv9-seg-dseg} & \textbf{YOLOv9-gelan-c-seg} & \textbf{YOLOv9-gelan-c-dseg} \\ \hline
\textbf{Mask mAP@50} & All & 0.59 & 0.579 & 0.576 \\ \hline
\textbf{Mask mAP@50} & Branch & 0.42 & 0.418 & 0.407 \\ \hline
\textbf{Mask mAP@50} & Trunk & 0.75 & 0.741 & 0.745 \\ \hline
\textbf{Box mAP@50} & All & 0.67 & 0.67 & 0.67 \\ \hline
\textbf{Box mAP@50} & Branch & 0.56 & 0.56 & 0.55 \\ \hline
\textbf{Box mAP@50} & Trunk & 0.78 & 0.78 & 0.78 \\ \hline
\end{tabular}
}
\end{table*}

\begin{figure}[h!]
    \centering
    \includegraphics[width=3.5 in, height=2 in]{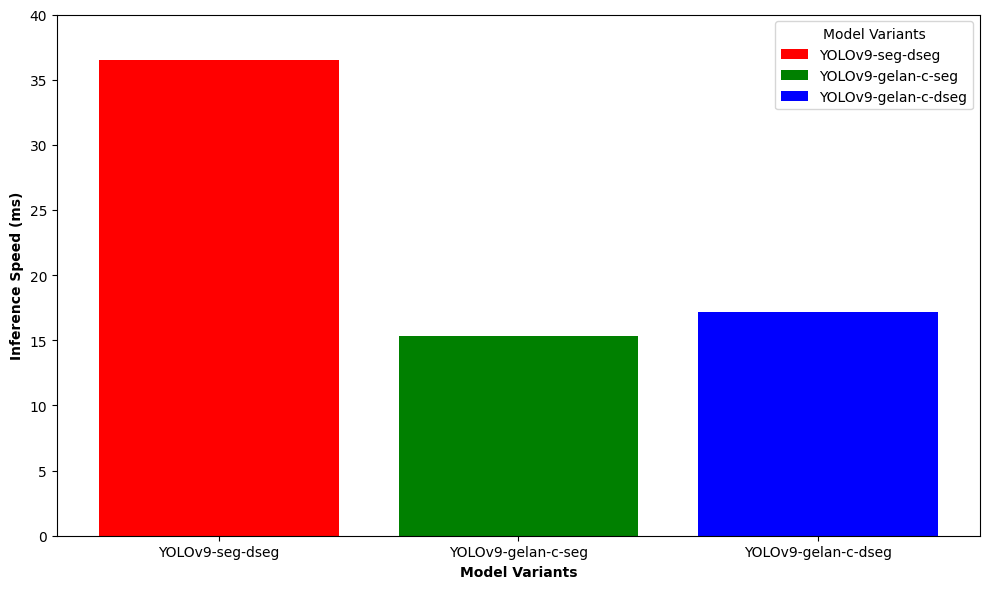}
    \caption{Inference Speed Comparison for YOLOv9 Models- showing the milliseconds required for trunk and branch segmentation by three YOLOv9 variants}
    \label{fig:FigureInferencespeed}
\end{figure}

Furthermore, the inference speed analysis revealed that the YOLOv9-gelan-c-seg model exhibited superior efficiency, achieving the fastest inference time of 15.3 ms, indicating a robust capacity for swiftly segmenting trunks and branches with high accuracy. Figure \ref{fig:FigureInferencespeed} shows the visualization of inference speeds for the three variants of YOLOv9 models for trunk and branch segmentation.  

\subsection{Kinect Fusion Reconstruction Validation}

\begin{figure*}[h!]
    \centering
    \includegraphics[width=6 in, height=6 in]{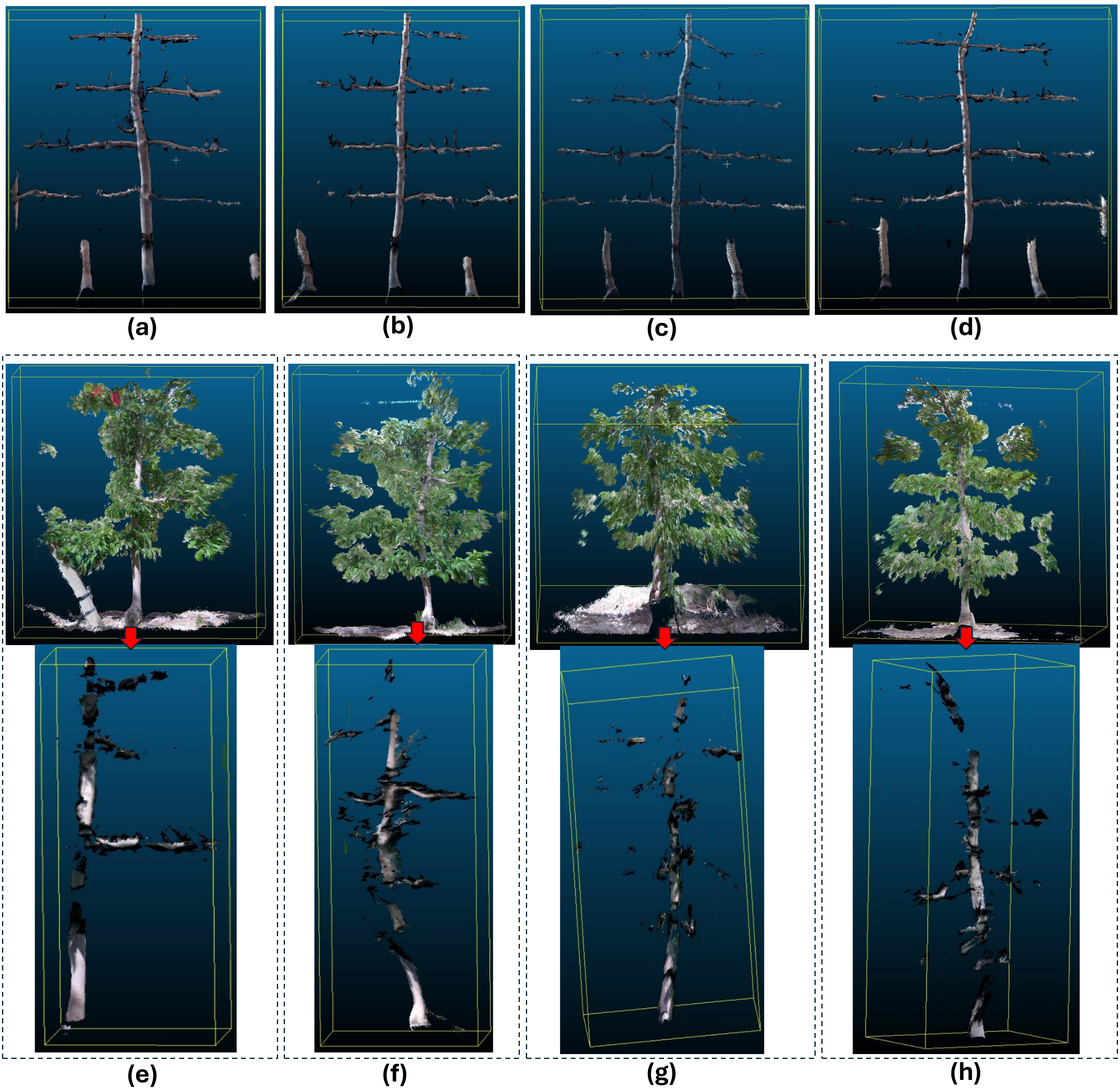}
    \caption{KinectFusion 3D Reconstruction Results Across Seasons: (a) Showcases a detailed 3D reconstruction of a dormant season apple tree, capturing trunks and branches post YOLOv9 segmentation, highlighting the model’s precision; (b-d) Additional examples from the dormant season further affirming the consistent and effective 3D modeling capabilities of KinectFusion; (e) Demonstrates trunk-focused 3D reconstruction in the canopy season due to dense foliage; (f-h) Subsequent canopy season images reveal significant branch data loss, with only the trunks prominently reconstructed, underscoring the challenges posed by dense foliage.}
    \label{fig:KinectFusion}
\end{figure*}

Figure \ref{fig:KinectFusion} showcases the KinectFusion 3D reconstructions derived from dormant and canopy season images following YOLOv9 segmentation. In the dormant season, a robust reconstruction of trunks and branches is evident, as demonstrated in Figure \ref{fig:KinectFusion}a, b, c, and d, where four examples of dormant season trees exhibit successful 3D reconstruction of detailed tree structures. This indicates the high accuracy and effectiveness of the KinectFusion method in capturing comprehensive structural details when the tree components are unobstructed. Conversely, the canopy season presents significant challenges, as illustrated in Figure \ref{fig:KinectFusion}e, f, g, and h. While trunks were generally segmented accurately, the reconstructions lacked substantial branch details. The upper portions of these figures display the original images, with the lower portions showing the post-segmentation KinectFusion reconstructions. It is clear that dense foliage severely impedes the segmentation process, resulting in a failure to capture and reconstruct a significant number of branches. Specifically, out of 109 branches for which imaging and ground truth data were collected, only 7 branches were partially reconstructed, showcasing the limitations of current segmentation and 3D reconstruction technologies under foliated conditions. The inability to effectively segment and reconstruct the majority of branches (102 out of 109) underscores the need for enhanced segmentation algorithms and reconstruction techniques that can penetrate dense canopy cover and accurately model the complex architecture of orchard trees during peak foliage periods.

The results from the recent analysis comparing root mean square error (RMSE) and mean absolute error (MAE) for trunk diameter, branch diameter, and branch spacing are documented in Table \ref{tab:measurement_metrics} and serve as a critical assessment of the precision achieved through the combined application of YOLOv9 segmentation and KinectFusion 3D reconstruction. Notably, the RMSE and MAE values for branch spacing significantly outperform those for trunk and branch diameters, suggesting superior accuracy in spacing measurements within the reconstructed 3D models. This phenomenon can be attributed to the fact that measuring spacing in a three-dimensional space generally requires fewer data points and is less susceptible to the complexities associated with the volumetric segmentation of tree components. 
\begin{table}[h]
\centering
\caption{Comparison of RMSE and MAE for Trunk Diameter, Branch Diameter, and Branch Spacing}
\label{tab:measurement_metrics}
\begin{tabular}{@{}lcc@{}}
\toprule
\textbf{Parameter} & \textbf{RMSE} & \textbf{MAE} \\ \midrule
Trunk Diameter & 5.233 & 4.683 \\
Branch Diameter & 4.50 & 3.22 \\
Branch Spacing & 0.54 & 0.48 \\ \bottomrule
\end{tabular}
\end{table} 

\begin{figure*}[h!]
    \centering
    \includegraphics[width=6 in, height=5 in]{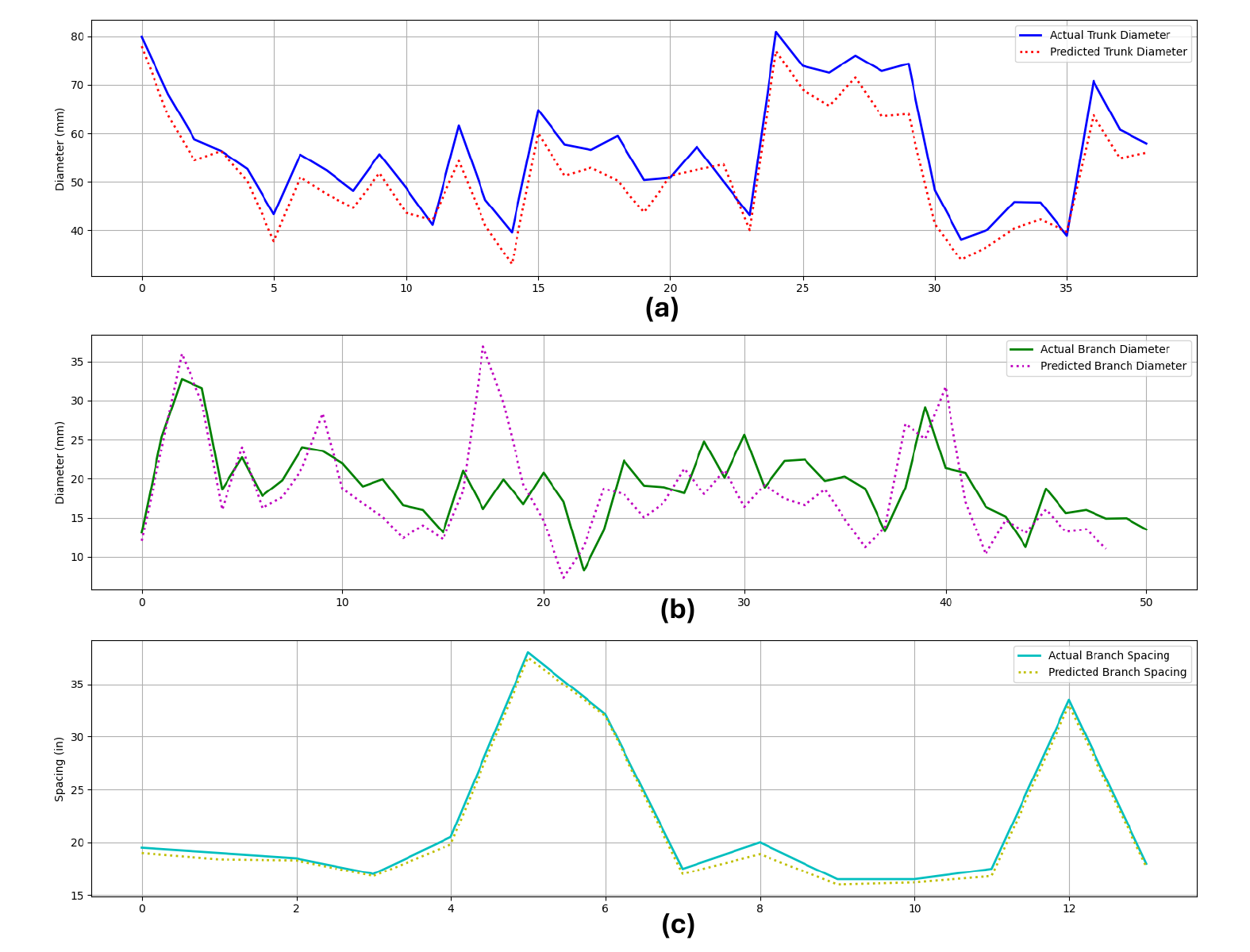}
    \caption{Comparative Analysis of Measurement Accuracy: (a) Illustrates the comparison of RMSE and MAE values for trunk diameter between reconstructed and ground-truth measurements; (b) Displays similar accuracy metrics for branch diameter; (c) Shows the precision in branch spacing measurements, highlighting the method's effectiveness in spatial accuracy.}
    \label{fig:KinectFusiongraph}
\end{figure*}
The RMSE for trunk diameter was recorded at 5.233 mm with an MAE of 4.683 mm, while branch diameter exhibited an RMSE of 4.50 mm and an MAE of 3.22 mm. These values highlight a commendable level of precision, particularly given the inherent challenges posed by the dense and overlapping foliage in canopy season images, which can obscure key structural features. Achieving such accuracy is pivotal for crop load management operations, where precise measurements of tree structures directly influence decision-making processes related to pruning, thinning, and overall orchard management. 

Furthermore, the branch spacing, recorded with an RMSE of 0.54 inches and an MAE of 0.48 inches, demonstrates exceptional accuracy, underscoring the efficacy of the segmentation and reconstruction approach in maintaining spatial integrity between tree elements. This level of detail is crucial for applications that rely on accurate spatial modeling to optimize the distribution of resources and light within the orchard. For a visual representation of these metrics, Figure \ref{fig:KinectFusiongraph} provides a comprehensive comparison of predicted versus actual measurements. Figure \ref{fig:KinectFusiongraph}a illustrates the comparison for trunk diameter, highlighting the proximity of reconstructed measurements to ground-truth data. Similarly, Figure \ref{fig:KinectFusiongraph}b demonstrates the alignment of predicted and actual measurements for branch diameter, while Figure \ref{fig:KinectFusiongraph}c focuses on branch spacing, further validating the reliability of the data produced by the 3D reconstruction process 

\subsection{Fast GICP Results}
\begin{figure*}[h!]
    \centering
    \includegraphics[width=7 in, height=6 in]{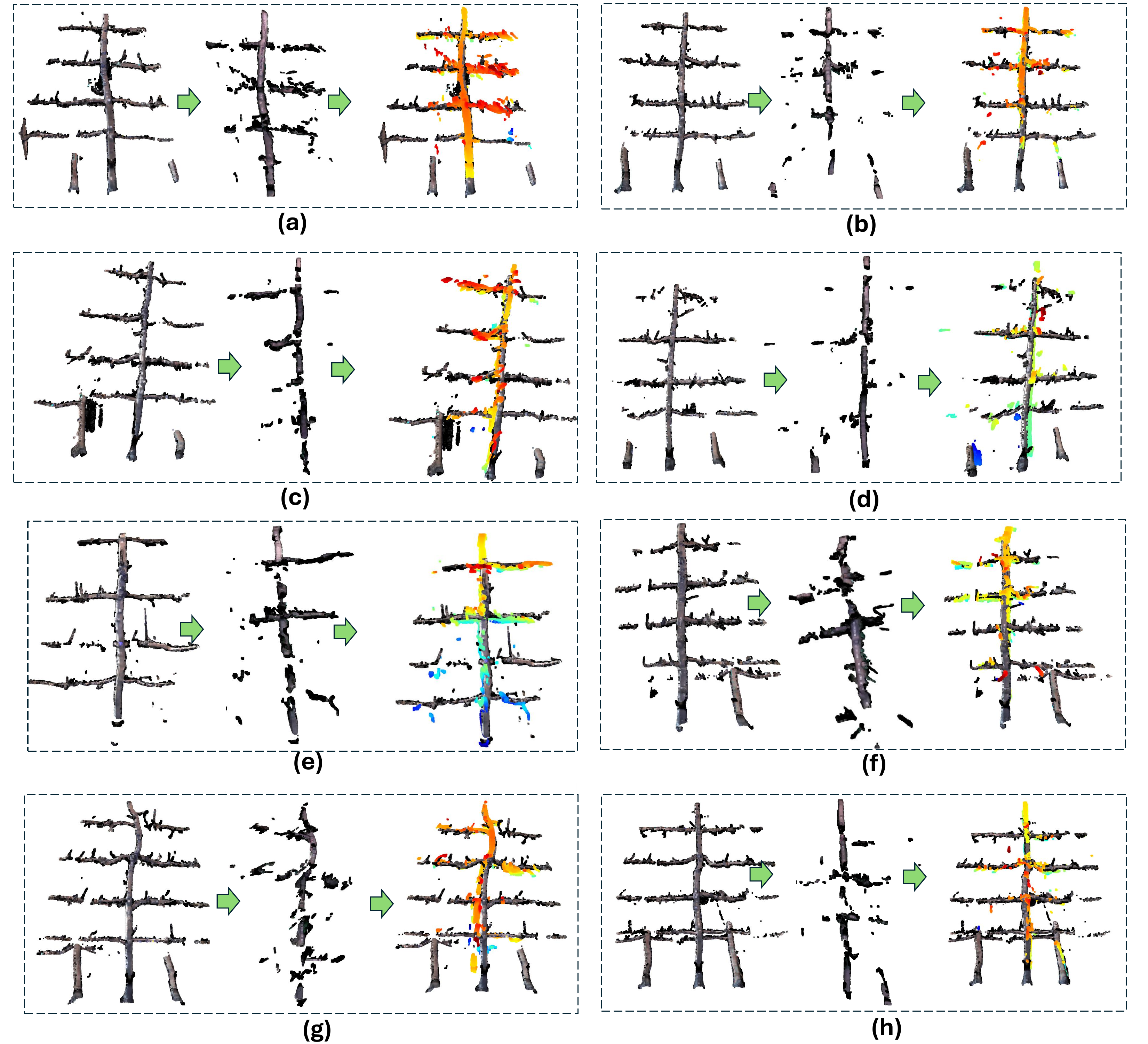}
    \caption{Figure illustrates Fast GICP registration of dormant-canopy season images post YOLOv9 segmentation and KinectFusion reconstruction. Subfigures (a-h) display sequences: leftmost images show segmented and reconstructed dormant season trees, middle images depict canopy season counterparts, and rightmost images present the total registration outcome, demonstrating the alignment effectiveness across seasonal variations.}
    \label{fig:gicpdiagram}
\end{figure*}

\begin{figure*}[h!]
    \centering
    \includegraphics[width=6 in, height=3.5 in]{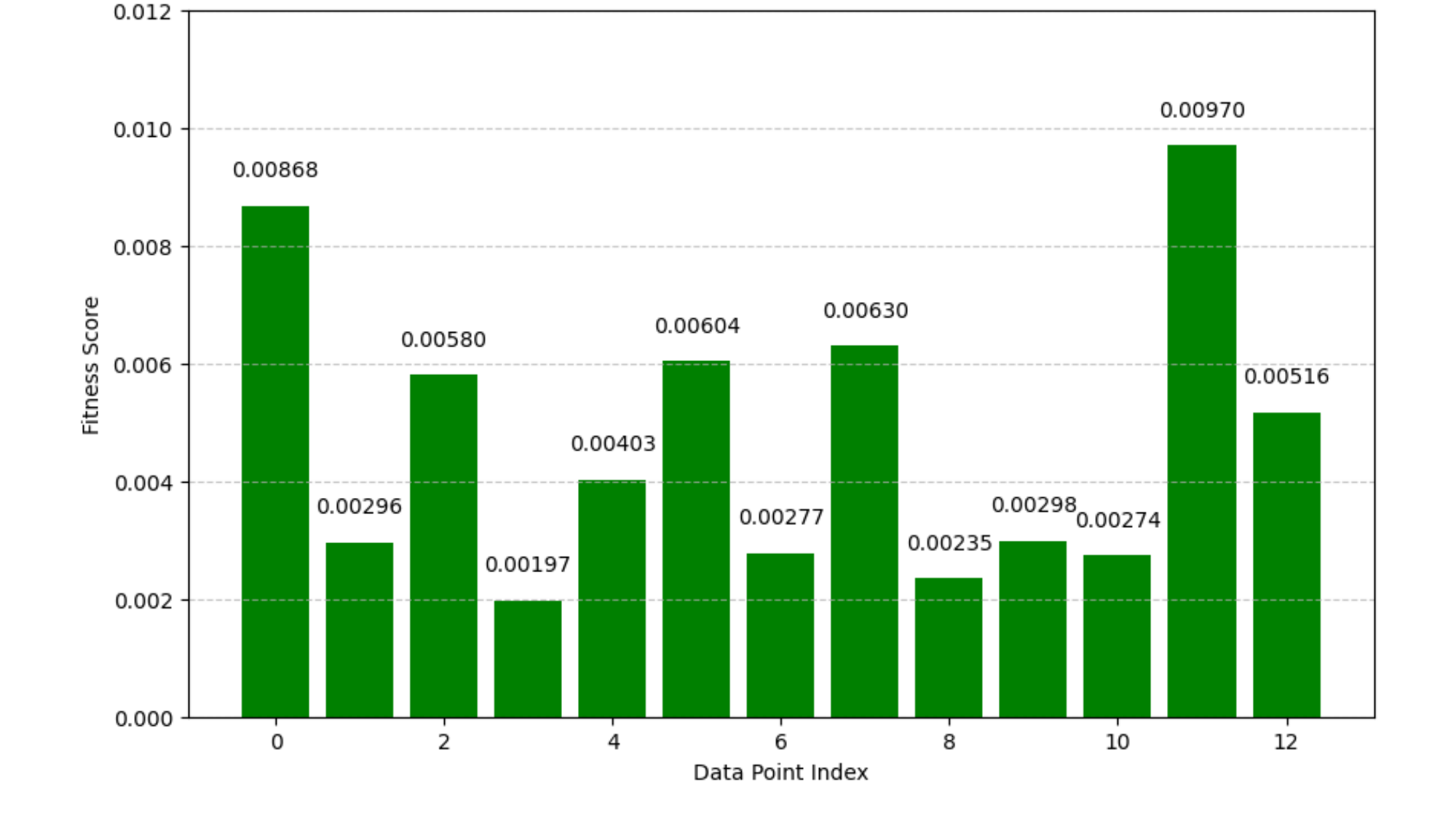}
    \caption{Bar diagram displaying Fast GICP fitness scores for seasonal image registration accuracy}
    \label{fig:fitnessscore}
\end{figure*}

The Fast GICP technique employed in this study demonstrated a promising ability to align and register 3D reconstructed point cloud data across different seasonal conditions specifically between the dormant and canopy seasons. This is effectively illustrated in Figure \ref{fig:gicpdiagram}, which showcases the comprehensive GICP registration process. In Figure \ref{fig:gicpdiagram}, each subfigure, labeled from (a) to (h), presents a visual triptych for eight different trees. The leftmost image in each set displays the canopy season's 3D reconstructed point cloud data, derived post YOLOv9 segmentation and KinectFusion 3D reconstruction, showcasing the complex tree structure amidst dense foliage. The middle image represents the corresponding dormant season data, where the lack of foliage provides a clearer view of the tree's structural framework. The rightmost image in each triptych illustrates the results of applying the Fast GICP technique, showing the final registered 3D point cloud that combines data from both seasons into a single, coherent model. This alignment process is critical for multiple reasons. First, it allows for a holistic analysis of tree growth and structural changes over different seasons by providing a unified 3D model that integrates visibility from both the leaf-off and leaf-on periods. Second, it enhances the precision of structural assessments and interventions such as pruning or disease management, by maintaining spatial accuracy across temporal transformations. The capability of Fast GICP to effectively merge disparate seasonal data sets into a singular, accurately aligned 3D representation not only confirms the robustness of the segmentation and reconstruction processes used but also opens up new avenues for continuous monitoring and management of orchard health and productivity.
The registration of dormant and canopy season images using thefFast GICP method yielded quantifiable results, as reflected by the analysis of the fitness scores (Figure \ref{fig:fitnessscore}. These scores represent the MSE between corresponding points in the aligned point clouds, offering a measure of the precision achieved in overlaying the 3D reconstructions from different seasonal captures. 
For the dormant season, the registration process demonstrated a higher degree of accuracy, with the best alignment recorded at an exceptionally low fitness score of 0.00197. This indicates an excellent alignment, where the average squared distance between matched points was minimal, showcasing the effectiveness of Fast GICP in handling clearer, less obstructed datasets typically found in dormant season imagery. Conversely, the canopy season images presented greater challenges in registration due to dense foliage, which often obscures critical structural details of the trees. The worst alignment within this dataset reached a fitness score of 0.00970, indicative of a relatively higher error margin. This result underscores the difficulty in achieving precise point cloud alignments under conditions of significant visual complexity and occlusion. Across all trials, the average fitness score was approximately 0.0047, indicating moderate alignment effectiveness across both seasons. This average suggests that while the Fast GICP method is capable of providing satisfactory alignment in many cases, there is variability in performance that likely stems from the inherent differences in image complexity between the dormant and canopy seasons. Despite the challenges posed by the canopy season images, most of the alignment scores remained below 0.006, a threshold generally deemed acceptable for robust 3D reconstruction and registration applications. This level of performance is promising for practical applications in orchard management, where accurate 3D spatial representations are crucial for decision-making processes such as pruning and disease management.

\section{Conclusion and Future Work}
innovative integration of 3D reconstruction and information fusion techniques using YOLOv9 and Fast GICP, aimed at enhancing structural analysis in apple orchards across dormant and canopy seasons. The effectiveness of the YOLOv9 segmentation model was significantly highlighted during the dormant season, where the visibility of tree structures allowed for precise delineation of trunks and branches. The clarity obtained during this season provided a robust dataset for subsequent 3D reconstruction and detailed structural analyses. Conversely, the canopy season presented considerable challenges, where dense foliage impeded effective segmentation, primarily restricting it to visible trunk parts,  where segmentation struggled against the complexity posed by the leaf-covered scenes. Despite these challenges, trunk detection was consistently successful, showcasing the model’s capability to recognize and delineate major tree structures even in less-than-ideal conditions. The application of Fast GICP facilitated the precise alignment and registration of 3D reconstructed models from both seasons, enhancing the ability to perform longitudinal studies and structural assessments across different seasonal phases. This alignment process was vital for merging the detailed dormant season data with the more obstructed canopy season imagery, thus providing a comprehensive year-round view of orchard dynamics. Validation efforts reinforced the reliability of these methodologies. Rigorous in-field measurements for trunk diameters, branch diameters, and branch spacings were compared against the model predictions, confirming the high accuracy of the 3D models generated from YOLOv9 segmented data. These validations were critical in establishing the practical utility of the segmentation and reconstruction techniques for real-world agricultural applications.

Building on the substantial advancements made in the field of orchard automation through the innovative use of 3D reconstruction and deep learning, the future of agricultural technology looks promising. The integration of these technologies has already demonstrated significant potential in enhancing the efficiency and precision of orchard management, particularly through the detailed structural analysis enabled by the fusion of dormant and canopy season imagery. As the industry moves forward, there are several key areas where automated and robotic operations could see substantial development and deployment. Firstly, the precision and accuracy achieved in the segmentation of tree structures during the dormant season suggest that similar methodologies could be expanded to include a variety of tree species and orchard configurations. This would allow for a broader application of the technology across different agricultural settings, potentially leading to widespread adoption in the sector.

Secondly, improving the capabilities of vision systems to operate effectively during the canopy season remains a critical challenge. Future research could focus on developing advanced imaging techniques and algorithms that enhance the system's ability to penetrate dense foliage and accurately identify crucial structural elements. This could include the use of multispectral imaging, LiDAR, or enhanced machine learning models that are better equipped to handle the complexities of leaf-covered scenes. Additionally, there is a significant opportunity to explore the automation of other labor-intensive tasks such as fruit picking, pest management, and disease detection. Robots equipped with the ability to not only recognize but also interact with the environment could drastically reduce the labor requirements and physical demands currently placed on human workers. These systems could be designed to operate autonomously or in conjunction with human operators, providing flexibility in how orchard management is approached.

Moreover, integrating these technologies with other smart farming solutions, such as IoT devices and precision agriculture tools, could create a highly interconnected system that manages various aspects of orchard operations. This would enable real-time data collection and analysis, leading to more informed decision-making and potentially higher yields and better quality produce. Finally, the development of robust models for predicting tree growth and fruit development based on the data collected from these automated systems could further enhance the strategic planning capabilities of orchard managers. By understanding the precise conditions and growth patterns of their orchards, managers could optimize resources, improve sustainability practices, and ultimately increase profitability. 

\section*{Acknowledgmenet and Funding}
This work was supported by Zhejiang Provincial Natural Science (Foundation Grant No. LD24E050006), The Zhejiang Provincial Key Research \& Development Program (Grant No. 2023C02049) the National Natural Science Foundation of China (Grant No. 32372004). Additionally, this work was supported by the National Science Foundation and the United States Department of Agriculture, National Institute of Food and Agriculture through the “Artificial Intelligence (AI) Institute for Agriculture” Program under Award AWD003473, and the AgAID Institute. 
%% The Appendices part is started with the command \appendix;
%% appendix sections are then done as normal sections

%% For citations use: 
%%       \cite{<label>} ==> [1]

%%

%% If you have bib database file and want bibtex to generate the
%% bibitems, please use
%%
%%  \bibliographystyle{elsarticle-num} 
%%  \bibliography{<your bibdatabase>}

%% else use the following coding to input the bibitems directly in the
%% TeX file.

%% Refer following link for more details about bibliography and citations.
%% https://en.wikibooks.org/wiki/LaTeX/Bibliography_Management

%\begin{thebibliography}{00}

\bibliographystyle{elsarticle-harv} % Use this for author-year
\bibliography{references}

\end{document}